\documentclass{article}

\date{} 

\usepackage{enumitem}

\usepackage{url}

\usepackage{hyperref} 

\usepackage{spconf,amsmath,graphicx,epstopdf}
\usepackage{multirow}
\usepackage{url}

\title{Multi-level Texture Encoding and Representation (MuLTER) based on Deep Neural Networks}
\name{Yuting Hu, Zhiling Long, and Ghassan AlRegib}
\address{Omni Lab for Intelligent Visual Engineering and Science (OLIVES)\\
Center for Signal and Information Processing (CSIP)\\
School of Electrical and Computer Engineering\\
Georgia Institute of Technology, Atlanta, GA 30332-0250, USA\\
\{huyuting, zhiling.long, alregib\}@gatech.edu}
\begin{document}
%
\onecolumn 

\begin{description}[labelindent=1cm,leftmargin=4cm,style=multiline]

\item[\textbf{Citation}]{Y. Hu, Z. Long, and G. AlRegib, ``Multi-level Texture Encoding and Representation (MuLTER) based on Deep Neural Networks,'' Proceedings of IEEE International Conference on Image Processing (ICIP), Sep. 2019.}
\\
\item[\textbf{Review}]{Date of acceptance: 30 April 2019}
\\
\item[\textbf{Data and Codes}]{\url{https://ghassanalregib.com/}}
\\
\item[\textbf{Bib}] {@INPROCEEDINGS\{multer, \\
author=\{Y. Hu and Z. Long and G. AlRegib\}, \\
booktitle=\{2019 IEEE International Conference on Image Processing (ICIP)\}, \\
title=\{Multi-level Texture Encoding and Representation (MuLTER) based on Deep Neural Networks\}, \\
year=\{2019\}, \\
month=\{Sep.\}\}
}
\\

\item[\textbf{Copyright}]{\textcopyright 2019 IEEE. Personal use of this material is permitted. Permission from IEEE must be obtained for all other uses, in any current or future media, including reprinting/republishing this material for advertising or promotional purposes, creating new collective works, for resale or redistribution to servers or lists, or reuse of any copyrighted component of this work in other works.}
\\
\item[\textbf{Contact}]{\href{mailto:alregib@gatech.edu}{alregib@gatech.edu}\\ \url{https://ghassanalregib.com/} \\ }
\end{description}

\thispagestyle{empty}
\newpage
\clearpage
\setcounter{page}{1}

\twocolumn

\maketitle
\begin{abstract}
In this paper, we propose a multi-level texture encoding and representation network (MuLTER) for texture-related applications. Based on a multi-level pooling architecture, the MuLTER network simultaneously leverages low- and high-level features to maintain both texture details and spatial information. Such a pooling architecture involves few extra parameters and keeps feature dimensions fixed despite of the changes of image sizes. In comparison with state-of-the-art texture descriptors, the MuLTER network yields higher recognition accuracy on typical texture datasets such as MINC-2500 and GTOS-mobile with a discriminative and compact representation. In addition, we analyze the impact of combining features from different levels, which supports our claim that the fusion of multi-level features efficiently enhances recognition performance. Our source code will be published on GitHub (https://github.com/olivesgatech).
\end{abstract}
\begin{keywords}
Texture encoding and representation, multi-level, convolutional neural network (CNN), texture pooling, feature fusion
\end{keywords}
\section{Introduction}
\label{sec:intro}

\begin{figure*}[htb]
  \centering
  \centerline{\includegraphics[width=17cm]{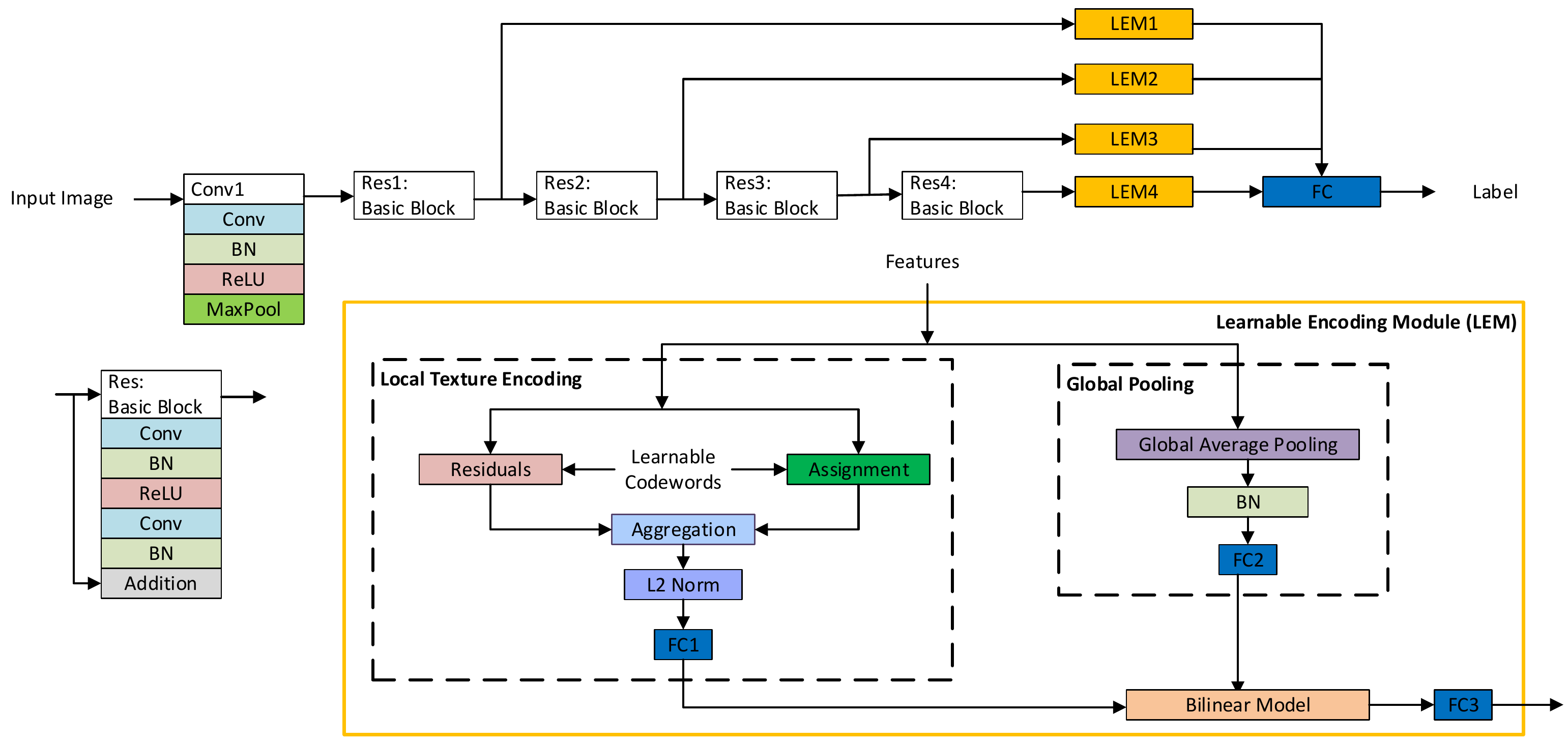}}
\caption{Flowchart of our proposed method where ``BN'' denotes batch normalization and ``FC'' represents a fully connected layer.}
\label{fig:flowchart}
\end{figure*}

Texture representation aims to extract descriptive features that provide important visual cues or characteristic object properties. ~\cite{liu2017local,cimpoi2015deep} provide a systematic review on widely used texture descriptors (both handcrafted and learning-based ones) and datasets. Nowadays, because of the record-breaking recognition accuracy, CNN~\cite{krizhevsky2012imagenet} has emerged as the new state-of-the-art tool for object recognition and classification. With a strong capability of preserving both local and global spatial information through convolutional, pooling, and nonlinear layers, CNN has become a universal representation for object recognition, in which spatial information is critical for identifying different objects. However, different from object recognition, texture and material recognition generally is challenging in demanding an orderless representation of micro-structures (i.e., texture encoding). Concatenated global CNN activations with a fully connected layer as a classifier have limitations in meeting the need for a geometry-invariant representation describing feature distributions.

Over the last decade, representation based on the bag-of-words (BOW) model has become a popular module in texture or object recognition and image understanding. BOW combined with local descriptors, such as the scale-invariant feature transform (SIFT)~\cite{lowe2004distinctive} or local binary patterns (LBP)~\cite{ojala1996comparative}, was the most widely used texture representation method. By assigning each local descriptor to its nearest visual word (i.e., a hard assignment), the BOW encoder calculates a histogram of visual word occurrences. To include richer information instead of simple occurrences, two popular extensions of BOW are vector of locally-aggregated descriptors (VLAD)~\cite{jegou2010aggregating} and Fisher vectors (FV)~\cite{perronnin2007fisher}. Different from BOW, VLAD accumulates the differences between a visual word and its corresponding local descriptors to aggregate first-order statistics of descriptors, while FV encodes both first- and second-order statistics of descriptors. Later, Cimpoi et al.~\cite{cimpoi2015deep} proposed a Fisher-vector CNN descriptor (FV-CNN), which is considered as a milestone for texture recognition with significantly boosted performance. FV-CNN computes FV pooling on generic deep features such as deep convolutional activation features (DeCAF). To generate these features, a deep CNN is pre-trained using the ImageNet~\cite{krizhevsky2012imagenet}. 
By removing the softmax and the last fully-connected layer of the network, DeCAF is then obtained as a feature vector. 

One shortcoming of the FV-CNN architecture is the separate learning of CNN feature extraction, texture encoding and classifier training, which does not benefit from the labeled data. To jointly learn them together in an end-to-end manner, Zhang et al.~\cite{zhang2017deep} proposed a texture encoding layer, which builds the dictionary learning and feature pooling on top of the CNN architecture. This deep texture encoding network (Deep-TEN) learns an orderless representation, which performs well on texture or material recognition. But as textures or materials do not always exhibit completely orderless patterns, local spatial information is still useful for differentiating them. To resolve this issue, Xue et al.~\cite{xue2018deep} presented a deep encoding pooling network (DEP), which fuses orderless texture encoding and local spatial information. However, neither Deep-TEN nor DEP fully utilizes CNN features from different layers and resolutions. Therefore, in this paper, we propose a multi-level convolutional neural network that improves over DEP. Our main contribution is an innovative network architecture that integrates features learned at different layers and embeds a learnable encoding module (LEM) at each individual layer. It extracts both low-level and high-level CNN features to achieve a multi-level texture representation, maintaining both texture details and local spatial information. 
The rest of the paper is organized as follows. Section 2 introduces the proposed texture descriptor MuLTER and its application in texture classification. Section 3 presents experimental results on popular texture datasets. Section 4 concludes.

\vspace{-0.2in}
\section{Proposed Method}
\label{sec:proposedmethod}

\begin{table*}[htb]
\centering
\caption{Architecture for adopting pretrained ResNet18.}
\resizebox{0.7\textwidth}{!}{
\begin{tabular}{|l|l|c|c|c|c|}
\hline
Modules & Layers                      &      Basic Blocks/Layers          & Output Size &  Multi-levels & LEM Output Size               \\ 
\hline
& Conv1                       & 7$\times$7, 64, stride 2                &       112$\times$112$\times$64           &  &     \\ 
\hline
\multirow{4}{*}{ResNet18} & Res1    & $\begin{bmatrix}
3\times 3,64  \\ 
3\times 3,64 
\end{bmatrix}\times 2$  &  56$\times$56$\times$64  & LEM1 & C=128  \\ 
\cline{2-6} 
& Res2                        &      $\begin{bmatrix}
3\times 3,128  \\ 
3\times 3,128 
\end{bmatrix}\times 2$            &        28$\times$28$\times$128       &  LEM2  &      C=128           \\ 
\cline{2-6}
&  Res3                        & $\begin{bmatrix}
3\times 3,256  \\ 
3\times 3,256 
\end{bmatrix}\times 2$                &   14$\times$14$\times$256        &  LEM3 &     C=128                  \\ 
\cline{2-6}
& Res4                        & $\begin{bmatrix}
3\times 3,512  \\ 
3\times 3,512 
\end{bmatrix}\times 2$                    &    7$\times$7$\times$512 &  LEM4   &           C=128            \\ 
\hline
Classifier & FC          &       128$\times$4 = 512 =\textgreater{}n               &            n classes        & &     \\ 
\hline
\end{tabular}
}
\label{tab:proposedarchitecture}
\end{table*}

To combine both low- and high-level CNN features, we propose a multi-level texture encoding and representation network (MuLTER), whose architecture is shown in Fig.~\ref{fig:flowchart} and Table~\ref{tab:proposedarchitecture}. We build the MuLTER on top of convolutional and non-linear layers pretrained on ImageNet~\cite{russakovsky2015imagenet} (e.g., ResNet18~\cite{he2016deep}). In addition, we incorporate modules of LEM at each individual layer.
\subsection{Learnable Encoding Module (LEM)}
\label{deepten}

\begin{table}[htb]
\caption{Learnable Encoding Module (LEM). The 3rd colum shows the output sizes for an input image size of $224\times224\times3$ and the 4th column shows the basic blocks or layers used.}
\begin{tabular}{|l|l|c|c|}
\hline
Spatial & Layers                      & Output size               & Basic Blocks/Layers                    \\
\hline
& Reshape                     & WH$\times$D                    & W$\times$H$\times$D =\textgreater{} WH$\times$D    \\ 

\hline
\multirow{2}{*}{Local} & Encoding                    & K$\times$D                     & K codewords                \\ 
\cline{2-4}

& Projection & 64 & FC1: KD =\textgreater{}64 \\ 
\hline

\multirow{2}{*}{Global} & Pooling                     & D                       & Average Pooling            \\ 
\cline{2-4}
& Projection                            & 64                        & FC2: 512 =\textgreater{} 64  \\ 
\cline{1-4}
L\&G & Bilinear              & 4096                      &        =\textgreater{}  $64^{2}$                   \\ 
\hline
& Projection                  & C=128                       & FC3: 4096 =\textgreater 128              \\ 

\hline
\end{tabular}
\label{tab:LEM}
\end{table}

For texture recognition in an end-to-end learning framework while maintaining texture details, the ``texture encoding'' layer was proposed~\cite{zhang2017deep}, which integrates dictionary learning and texture encoding in a single learnable model on top of convolutional layers, shown in Fig.~\ref{fig:flowchart}. It learns an inherent dictionary of local texture descriptors extracted from CNNs and generalizes robust residual encoders such as VLAD~\cite{jegou2010aggregating} and Fisher Vector~\cite{perronnin2010improving} through a ``residual'' layer calculated by pairwise difference between texture descriptors and the codewords of the dictionary. In ``assignment'' layer, assignment weights are calculated based on pairwise distance between texture descriptors and codewords and the ``aggregation'' layer converts the residuals vectors and the assignment weights into a full image representation. Thanks to the residual encoding, such image representations discarding frequently appearing features are helpful to domain transfer learning.

In addition to orderless texture details captured by the encoding layer, local spatial information are important visual cues, and the ``global pooling'' layer~\cite{xue2018deep} preserves local spatial information by average pooling. Then, a bilinear model~\cite{freeman1997learning} follows the texture encoding layer and the global pooling layer to jointly combine the two types of complementary information. We refer to the entire module as a learnable encoding module (LEM), shown in Fig.~\ref{fig:flowchart} and Table~\ref{tab:LEM}. Here we briefly introduce the notations. The input size of a LEM is $W\times H\times D$, where $W$, $H$, and $D$ denote the width, the height, and the feature channel dimension of the input volume, respectively. The codewords' number of the learnable dictionary is $K$. 

\subsection{Multi-level Deep Feature Fusion}
\label{subsec:multi-levelfeaturefusion}

The multi-level feature fusion means the joint utilization of both low-level features and high-level features from Res1 to Res4 of ResNet18. ResNet18 uses 4 basic blocks of similar structures and one example of the basic block is shown in the left bottom of Fig.~\ref{fig:flowchart}. Given an input image with size $224\times224\times3$, after employing convolutional filters (i.e. Conv1, a default structure at the beginning of the Resnet family), the output size is $112\times112\times64$. Then we feed it into ResNet18. Here, we have four levels, Res1, Res2, Res3, and Res4.   

The outputs from each level have different output sizes so we feed them into different sizes of LEMs. For example, for the first level, Res1 is followed by LEM1, where the output size of Res1 is $W\times H\times D=112\times 112\times 64$ and LEM1 converts it into a feature vector of dimension $C=128$. Whatever the input image size is, the same architecture shown in Table~\ref{tab:proposedarchitecture} can be used to produce a fixed-length (i.e., $C$) feature representation. Similar to the first level, we can repeat the procedure above to calculate a feature vector of dimension $C=128$ for level 2, 3, and 4. For local CNN-based texture descriptors at each level with either low-level features or high-level features, we preserve both texture details and local spatial information through their corresponding LEMs. To combine the features from different levels, we concatenate them and feed them into a classification layer. Assuming the number of classes is $n$, the classification layer maps the $4C$ feature vector to $n$ classes.

The multi-level architecture for texture encoding and representation has multiple advantages. First, the multi-level architecture makes it easy to adjust regarding which level of information should be fused. Second, it can be easily extended to other CNN models (e.g. ResNet50) by adapting the size of LEMs and the number of levels. Third, all modules in the overall architecture are differentiable, so the network can be trained with back propagation in an end-to-end texture encoding and representation network. Last but not the least, this architecture produces a compact yet discriminative representation with a full image representation with a dimension of a few hundreds (e.g. 512).

\section{Experiments}
\label{sec:experiments}

\subsection{Datasets and Implementation Details}
\label{subsec: datasets}
\textbf{Datasets}: to show the recognition performance of our proposed method for texture representation, we test it on two recent challenging texture or material datasets: materials in context database (MINC)-2500~\cite{bell2015material} and ground terrain database (GTOS)-mobile~\cite{xue2018deep}. The MINC dataset is an order of magnitude larger than previous texture and material datasets (such as KTH-TIPS~\cite{caputo2005class} and FMD~\cite{sharan2009material}), while being more diverse and well-sampled across its 23 categories. For a fair comparison with other methods, we use MINC-2500 (i.e. a subset of MINC with 2500 patches per category). GTOS-mobile is a dataset including images for ground terrain regions captured by mobile phones. It consists of 31 classes such as grass, brick, soil, etc., and can be used for material classification. The GTOS-mobile is challenging because of its realistic capturing conditions (i.e. a mobile imaging device, handheld video, and uncalibrated capture). Compared with GTOS-mobile, MINC-2500 is a more general one.

\textbf{Implementations}: following the standard testing protocol of MINC-2500 and GTOS-mobile, we use the same data augmentation and training procedure as in~\cite{xue2018deep}. We resize images to $256\times 256$ and randomly crop patches to $224\times224$. For the training part, we augment data using horizontal flips with a 50$\%$ probability. For a fair comparison with~\cite{xue2018deep}, we build a ResNet18 for the GTOS-mobile dataset and build a ResNet50 for the MINC-2500 dataset. As Sec.~\ref{subsec:multi-levelfeaturefusion} mentions, our method is easily extended to other CNN models (e.g. ResNet50) by adapting the size of LEMs. Our experimental settings are: learning rate starting at 0.01 and decaying every 10 epochs by a factor of 0.1, batch size 128 for GTOS-mobile and 32 for MINC-2500, momentum 0.9, and the total number of epochs 30. The number of codewords $K$ is set to 8 for GTOS-mobile and 32 for MINC-2500. The result is shown in Table~\ref{tab: comparasionresults}, which shows the superior recognition accuracy of our proposed multi-level architecture. We run experiments on a PC (Nvidia GeForce GTX1070, RAM: 8GB).  

\label{subsec: results}
\begin{table}[t]
\begin{center}
\caption{Comparison of various level-selection schemes of our proposed method on the MINC-2500 and the GTOS-mobile datasets.}
\label{tab: proposed_schemes}
\resizebox{0.40\textwidth}{!}{
    \begin{tabular}{|c|c|c|}
    \hline
    Schemes      & MINC-2500~\cite{bell2015material} & GTOS-mobile~\cite{xue2018deep} \\
    \hline
    L=1 &  59.10\%    &  62.35\%  \\
    \hline
    L=2 & 70.84\%     &  74.94\%\\
    \hline
    L=3 & 80.70\%     &  76.43\% \\
    \hline
    L=4 & 81.01\%     &   77.04\% \\
    \hline
    L=1,2 &  70.16\%      & 77.68\%   \\
    \hline
    L=3,4 &   81.29\%     & 76.08\% \\
    \hline
    L=1,4 &   81.29\%     & 77.88\%   \\
    \hline
    L=1,2,3 &   80.45\%     & 75.44\%   \\
    \hline
    L=2,3,4 &    81.44\%     & 76.46\%   \\
    \hline
    L=1,2,3,4 &  82.21\%       & 78.21\%    \\
    \hline
    \end{tabular}
}
\end{center}
\end{table}

\subsection{Results}
\textbf{Impact of Level Selections}: Table~\ref{tab: proposed_schemes} shows the results obtained from various schemes of level selection on MINC-2500 and GTOS-mobile, separately. Each scheme utilizes CNN features of different levels, from single levels (e.g. L=1 or L=4) to multiple levels (e.g. L=1,4 or L=1,2,3,4). Table~\ref{tab: proposed_schemes} does not include an exhaustive comparison of different schemes since we skip those similar results and show the representative ones. From Table~\ref{tab: proposed_schemes}, we have several observations:

(1) For single levels on both datasets, the results under setting ``L=4'' outperforms setting ``L=1''. This indicates that with the CNN architectures, high-level features tend to describe textures better than low-level features. Setting ``L=4'' performs 21.91$\%$ better than ``L=1'' on the MINC-2500 dataset and 14.69$\%$ on the GTOS-mobile dataset.

(2) The setting ``L=1,2,3,4'' with four levels achieves the highest recognition accuracy on the MINC-2500 dataset, and yields the highest accuracy on GTOS-mobile dataset. This observation supports our claim that fusing information from multiple levels improves the discriminative capability to better describe and differentiate various texture images.

(3) The benefits from multi-level feature fusion vary among datasets. For example, on the GTOS-mobile dataset, the second highest recognition accuracy obtained from setting ``L=1,4'' is just 0.33$\%$ lower than that of setting ``L=1,2,3,4'', which implies that ``L=1,4'' already captures sufficiently discriminative features, while features from``L=2,3'' bring limited improvement. In contrast, on the MINC-2500 dataset, the second highest recognition accuracy comes from setting ``L=2,3,4'', 0.15$\%$ higher than that of setting ``L=1,4'' without medium-level features. Thus, features from setting``L=1'' or ``L=2,3'' have comparable contributions to the improvement of the discriminative capability, but which one brings more improvement depends on specific datasets.

(4) The impact of incorporating features from certain levels can vary significantly among different datasets. For example, on the GTOS-mobile dataset, setting ``L=1,2'' already yields good performance and outperforms that from setting ``L=3,4'' while the MINC-2500 datset presents the opposite results. This indicates that for a larger, more diverse dataset like MINC-2500, features from deeper layers bring more improvement than those from shallow layers.

\textbf{Comparison with State-of-the-art Methods}: we evaluated our method and compared with other state-of-the-art methods on the datasets mentioned above. The results for ResNet~\cite{he2016deep}, FV-CNN~\cite{cimpoi2014describing}, and Deep-TEN~\cite{zhang2017deep} were borrowed from~\cite{xue2018deep}. The results for DEP were generated using codes~\cite{dep_github} provided by the authors. On the MINC-2500 dataset, our method achieved a recognition accuracy of 82.2\%, which outperforms Deep-TEN by 1.8$\%$ and DEP by 1.2$\%$. On the GTOS-mobile dataset, the recognition accuracy of our method is 77.9$\%$, which is 4.0$\%$ better than Deep-TEN and 1.2$\%$ better than DEP. The reason behind our superior performance is that our method fuses multi-level CNN features in a distinctive and compact way while other methods only use features from a single level (i.e., the last layer).

\begin{table}[t]
\begin{center}
\caption{Comparison with state-of-the-art algorithms on the MINC-2500 and the GTOS-mobile datasets.}
\label{tab: comparasionresults}
\resizebox{0.42\textwidth}{!}{
    \begin{tabular}{|c|c|c|c|}
    \hline
    Method     &  MINC-2500~\cite{bell2015material} & GTOS-mobile~\cite{xue2018deep} \\
    \hline
    ResNet~\cite{cimpoi2015deep}    & N/A        &     70.8\% \\
    \hline
    FV-CNN~\cite{cimpoi2015deep}      & 63.1\%        &     N/A    \\         \hline 
    Deep-TEN~\cite{zhang2017deep}  & 80.4\% 
    & 74.2\%         
    \\
    \hline
    DEP~\cite{xue2018deep}  &  81.0\%                                  
    & 77.0\%     
    \\
    \hline
    \textbf{Proposed}  & 82.2\%      
    & 78.2\%     
    \\
    \hline
    \end{tabular}
}
\end{center}
\end{table}

\section{Conclusion}
\label{sec:conclusion}

\vspace{-0.15in}

We proposed a multi-level deep architecture (MuLTER) in this paper. It fulfilled a multi-level texture representation, simultaneously extracting low-level and high-level CNN features to maintain texture details and local spatial information. In comparison with the state-of-the-art techniques, MuLTER has accomplished higher recognition accuracy with a compact feature representation on two challenging texture datasets. Additionally, we analyzed the impact of incorporating CNN features from different levels on our proposed method.




\bibliographystyle{IEEEbib}
\bibliography{main}

\end{document}